\pdfoutput=1
\documentclass[11pt,letterpaper]{article}
\usepackage{emnlp2017}
\usepackage{times}
\usepackage{latexsym}

\usepackage{amsmath}
\usepackage{multirow}
\usepackage{url}
\makeatletter

\usepackage{amsfonts}
\usepackage{tikz}
\usetikzlibrary{arrows,shapes,trees}

\emnlpfinalcopy

\def\emnlppaperid{437}

\title{Sentence Simplification with Deep Reinforcement Learning}

\date{}

\author{Xingxing Zhang \and Mirella Lapata \\
	Institute for Language, Cognition and Computation
	\\
	School of Informatics, University of Edinburgh
	\\ 
	10 Crichton Street, Edinburgh EH8 9AB \\
	{\tt x.zhang@ed.ac.uk,mlap@inf.ed.ac.uk}
}

\date{}

\begin{document}

\maketitle

\begin{abstract}
  Sentence simplification aims to make sentences easier to read and
  understand. Most recent approaches draw on insights from machine
  translation to learn simplification rewrites from monolingual
  corpora of complex and simple sentences. We address the
  simplification problem with an encoder-decoder model coupled with a
  deep reinforcement learning framework. Our model, which we call {\sc
    Dress} (as shorthand for {\bf D}eep {\bf RE}inforcement {\bf
    S}entence {\bf S}implification), explores the space of possible
  simplifications while learning to optimize a reward function that
  encourages outputs which are simple, fluent, and preserve the
  meaning of the input. Experiments on three datasets demonstrate that
  our model outperforms competitive simplification
  systems.\footnote{Our code and data are publicly available at
    \url{https://github.com/XingxingZhang/dress}.}
	
\end{abstract}

\section{Introduction}
\label{sec:introduction}
The main goal of \emph{sentence simplification} is to reduce the
linguistic complexity of text, while still retaining its original
information and meaning.  The simplification task has been the subject
of several modeling efforts in recent years due to its relevance for
NLP applications and individuals alike
\cite{Siddharthan:2014,Shardlow:2014}. For instance, a simplification
component could be used as a preprocessing step to improve the
performance of parsers \cite{Chandrasekar:ea:96}, summarizers
\cite{Klebanov:ea:04}, and semantic role labelers
\cite{vickrey-koller:2008:ACLMain,Woodsend:Lapata:2014}. Automatic
simplification would also benefit people with low-literacy skills
\cite{Watanabe:ea:09}, such as children and non-native speakers as
well as individuals with autism \cite{Evans:ea:2014}, aphasia
\cite{Carroll:ea:99}, or dyslexia \cite{Rello:ea:2013}.

The most prevalent rewrite operations which give rise to simplified
text include substituting rare words with more common words or
phrases, rendering syntactically complex structures simpler, and
deleting elements of the original text
\cite{Siddharthan:2014}. Earlier work focused on individual aspects of
the simplification problem. For example, several systems performed
syntactic simplification only, using rules aimed at sentence splitting
\citep{Carroll:ea:99,Chandrasekar:ea:96,vickrey-koller:2008:ACLMain,Siddharthan:06}
while others turned to lexical simplification by substituting
difficult words with more common WordNet synonyms or paraphrases
\citep{Devlin:1999,inui-EtAl:2003:PARAPHRASE,kaji-EtAl:2002:ACL}.

Recent approaches view the simplification process more holistically as
a monolingual text-to-text generation task borrowing ideas from
statistical machine translation. Simplification rewrites are learned
automatically from examples of complex-simple sentences extracted from
online resources such as the ordinary and simple English
Wikipedia. For example, \newcite{zhu2010monolingual} draw inspiration
from syntax-based translation and propose a model similar to
\newcite{yamada2001syntax} which additionally performs
simplification-specific rewrite operations (e.g., sentence
splitting). \newcite{woodsend-lapata:2011:EMNLP} formulate
simplification in the framework of Quasi-synchronous grammar
\cite{smith2006quasi} and use integer linear programming to score the
candidate translations/simplifications. \newcite{wubben2012sentence}
propose a two-stage model: initially, a standard phrase-based machine
translation (PBMT) model is trained on complex-simple sentence
pairs. During inference, the $K$-best outputs of the PBMT model are
reranked according to their dis-similarity to the (complex) input
sentence. The hybrid model developed in \newcite{narayan-gardent:2014}
also operates in two phases.  Initially, a probabilistic model
performs sentence splitting and deletion operations over discourse
representation structures assigned by Boxer
\cite{curran-clark-bos:2007:PosterDemo}.  The resulting sentences are
further simplified by a model similar to \newcite{wubben2012sentence}.
\newcite{Xu_TACL16} train a syntax-based machine translation model on
a large scale paraphrase dataset \cite{ganitkevitch2013ppdb} using
simplification-specific objective functions and features to encourage
simpler output.

In this paper we propose a simplification model which draws on
insights from neural machine translation
\cite{bahdanau2014neural,sutskever2014sequence}.  Central to this
approach is an encoder-decoder architecture implemented by recurrent
neural networks.  The encoder reads the source sequence into a list of
continuous-space representations from which the decoder generates the
target sequence. Although our model uses the encoder-decoder
architecture as its backbone, it must also meet constraints imposed by
the simplification task itself, i.e.,~the predicted output must be
simpler, preserve the meaning of the input, and grammatical. To
incorporate this knowledge, the model is trained in a reinforcement
learning framework \cite{williams1992simple}: it explores the space of
possible simplifications while learning to maximize an expected reward
function that encourages outputs which meet simplification-specific
constraints. Reinforcement learning has been previously applied to
extractive summarization \cite{ryang-abekawa:2012:EMNLP}, information
extraction \cite{narasimhan:2016:EMNLP2016}, dialogue generation
\cite{li-EtAl:2016:EMNLP}, machine translation, and image caption
generation \cite{ranzato2016sequence}.

We evaluate our system on three publicly available datasets collated
automatically from Wikipedia
\cite{zhu2010monolingual,woodsend-lapata:2011:EMNLP} and
human-authored news articles \cite{Xu-EtAl:2015:TACL}. We
experimentally show that the reinforcement learning framework is the key
to successful generation of simplified text bringing significant
improvements over strong simplification models across datasets. 

\section{Neural Encoder-Decoder Model}
\label{sec:encdeca}

We will first define a basic encoder-decoder model for sentence
simplification and then explain how to embed it in a reinforcement
learning framework. Given a (complex) \emph{source} sentence
\mbox{$X = (x_1, x_2, \dots, x_{|X|})$}, our model learns to predict
its simplified \emph{target} $Y = (y_1, y_2, \dots,
y_{|Y|})$.
Inferring the target~$Y$ given the source $X$ is a typical sequence to
sequence learning problem, which can be modeled with attention-based
encoder-decoder models
\cite{bahdanau2014neural,luong-etal:2015:EMNLP}. Sentence
simplification is slightly different from related sequence
transduction tasks (e.g.,~compression) in that it can involve
splitting operations. For example, a long source sentence (\textsl{In
	1883, Faur married Marie Fremiet, with whom he had two sons.}) can
be simplified as two sentences (\textsl{In 1883, Faur married Marie
	Fremiet. They had two sons.}).  Nevertheless, we still view the
target as a sequence, i.e.,~two or more sequences concatenated with
full stops.

\begin{figure*}
	
	\centering
	\resizebox{\textwidth}{!}{%
		
		\centering
		\begin{tikzpicture}[scale=.75,->,>=stealth',thick,main node/.style={rectangle,rounded corners=3pt,fill=blue!10,draw,font=\sffamily\Large\bfseries,inner sep=0,minimum size=2.5mm,minimum width=3mm,minimum height=1cm,path picture={
				\draw[fill=blue!50!black] (0, -0.25) circle (0.8mm);
				\draw[fill=blue!50!black] (0, 0.25) circle (0.8mm);
			}
		},
		word node/.style={rectangle,rounded corners=3pt,fill=red!30!white,draw,font=\sffamily\bfseries,inner sep=3pt,minimum size=2.5mm}
		]
		
		\node[main node] (xh_1) at (-10, 0) {};
		\node[main node] (xh_2) at (-8, 0) {};
		\node[main node] (xh_3) at (-6, 0) {};
		\node[main node] (xh_4) at (-4, 0) {};
		\node[main node] (xh_5) at (-2, 0) {};
		
		\draw[line width=2pt,blue] (-12, 0) -- (xh_1);
		\draw[line width=2pt,blue] (xh_1) -- (xh_2);
		\draw[line width=2pt,blue] (xh_2) -- (xh_3);
		\draw[line width=2pt,blue] (xh_3) -- (xh_4);
		\draw[line width=2pt,blue] (xh_4) -- (xh_5);
		
		\node (x) at (-11.5, -2) {\Large $X=$};
		\node (x1) at (-10, -2) {\Large $x_1$};
		\node (x2) at (-8, -2) {\Large $x_2$};
		\node (x3) at (-6, -2) {\Large $x_3$};
		\node (x4) at (-4, -2) {\Large $x_4$};
		\node (x5) at (-2, -2) {\Large $x_5$};
		
		\draw[line width=2pt,green] (x1) -- (xh_1);
		\draw[line width=2pt,green] (x2) -- (xh_2);
		\draw[line width=2pt,green] (x3) -- (xh_3);
		\draw[line width=2pt,green] (x4) -- (xh_4);
		\draw[line width=2pt,green] (x5) -- (xh_5);
		
		\node[main node] (c) at (-4, 2.5) {};
		
		\draw[line width=1.5pt,red] (-10, 0.7) -- (-4, 1.8);
		\draw[line width=1.5pt,red] (-8, 0.7) -- (-4, 1.8);
		\draw[line width=1.5pt,red] (-6, 0.7) -- (-4, 1.8);
		\draw[line width=1.5pt,red] (-4, 0.7) -- (-4, 1.8);
		\draw[line width=1.5pt,red] (-2, 0.7) -- (-4, 1.8);
		
		\draw[line width=1pt,black,dashed] (-6, 2.4) -- (-10, 0.7);
		\draw[line width=1pt,black,dashed] (-6, 2.4) -- (-8, 0.7);
		\draw[line width=1pt,black,dashed] (-6, 2.4) -- (-6, 0.7);
		\draw[line width=1pt,black,dashed] (-6, 2.4) -- (-4, 0.7);
		\draw[line width=1pt,black,dashed] (-6, 2.4) -- (-2, 0.7);
		
		\node[main node] (yh_1) at (-10, 3) {};
		\node[main node] (yh_2) at (-8, 3) {};
		\node[main node] (yh_3) at (-6, 3) {};
		
		\draw[line width=2pt,black] (-12, 3) -- (yh_1);
		\draw[line width=2pt,black] (yh_1) -- (yh_2);
		\draw[line width=2pt,black] (yh_2) -- (yh_3);
		
		\node (y1) at (-11.5, 5) {\Large $\hat{Y} = $};
		\node (y1) at (-10, 5) {\Large $\hat{y}_1$};
		\node (y2) at (-8, 5) {\Large $\hat{y}_2$};
		\node (y3) at (-6, 5) {\Large $\hat{y}_3$};
		
		\draw[line width=2pt,black] (yh_1) -- (y1);
		\draw[line width=1.5pt,green] (y1) -- (yh_2);
		\draw[line width=2pt,black] (yh_2) -- (y2);
		\draw[line width=1.5pt,green] (y2) -- (yh_3);
		\draw[line width=2pt,black] (yh_3) -- (y3);
		\draw[line width=2pt,black] (c) -- (-4, 4.5) -- (-5.8, 4.5);
		
		
		\draw[line width=4pt,black] (-1, 3) -- node[above = 10pt] { {\bf Get Action Seq.} $\hat{Y}$ } (1, 3);
		
		\draw[line width=4pt,black] (1, -0.5) -- node[above = 10pt] { {\bf Update Agent} } (-1, -0.5);
		
		\node[word node, align = center, fill=blue!25!white] (r_simplicity) at (4, 3) {Simplicity \\ Model};
		
		\node[word node, align = center, fill=red!25!white] (r_relevance) at (7, 3) {Relevance \\ Model};
		
		\node[word node, align = center, fill=yellow!20!white] (r_grammar) at (10, 3) {Grammar \\ Model};
		
		\node[word node, align = center, fill=purple!20!green!20] (reinforce) at (7, -0.5) {\LARGE REINFORCE algorithm};
		
		\draw[line width=4pt,black] (4, 2) -- (4, 0.5);
		\draw[line width=4pt,black] (7, 2) -- (7, 0.5);
		\draw[line width=4pt,black] (10, 2) -- (10, 0.5);
		
		\node (hat_Y_G) at (10, 5) {\Large $\hat{Y}$};
		\draw[line width=2pt,green] (hat_Y_G) -- (r_grammar);
		
		\node (X_R) at (6, 5) {\Large $X$};
		\node (hat_Y_R) at (8, 5) {\Large $\hat{Y}$};
		\draw[line width=2pt,green] (X_R) -- (r_relevance);
		\draw[line width=2pt,green] (hat_Y_R) -- (r_relevance);
		
		\node (X_S) at (3, 5) {\Large $X$};
		\node (hat_Y_S) at (4, 5) {\Large $\hat{Y}$};
		\node (Y_S) at (5, 5) {\Large $Y$};
		\draw[line width=2pt,green] (X_S) -- (r_simplicity);
		\draw[line width=2pt,green] (hat_Y_S) -- (r_simplicity);
		\draw[line width=2pt,green] (Y_S) -- (r_simplicity);
		
		\end{tikzpicture}
		
	}%
	\caption{\label{fig:model-overview} Deep reinforcement
		learning simplification model. $X$ is the complex sentence,
		$Y$ the reference (simple) sentence and
		$\hat{Y}$ the action
		sequence (simplification) produced by the
		encoder-decoder model.}
	\label{fig:sys}
\end{figure*}
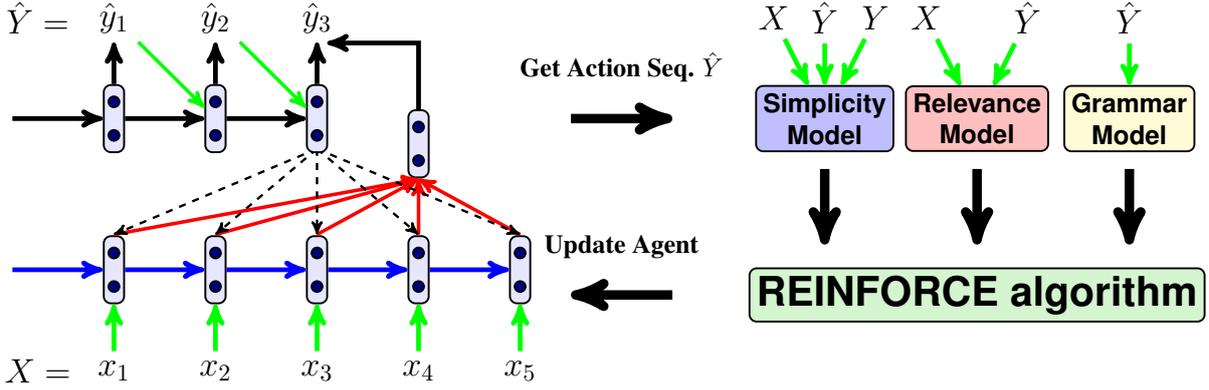

The encoder-decoder model has two parts (see left hand side in
Figure~\ref{fig:sys}). The \emph{encoder} transforms the source
sentence $X$ into a sequence of hidden states
$(\mathbf{h}^S_1, \mathbf{h}^S_2, \dots, \mathbf{h}^S_{|X|})$ with a
Long Short-Term Memory Network (LSTM; \citealt{hochreiter1997long}),
while the \emph{decoder} uses another LSTM to generate one word
$y_{t+1}$ at a time in the simplified target~$Y$.
Generation is conditioned on all previously generated words
$y_{1:t}$ and a dynamically created context vector $\mathbf{c}_t$,
which encodes the source sentence:
\begin{equation}
\label{eq:sent_prob}
P(Y | X) = \prod_{t=1}^{|Y|} P(y_t|y_{1:t-1},X)
\end{equation}
\begin{equation}
\label{eq:policy}
P(y_{t+1}|y_{1:t}, X) = \text{softmax}( g( \mathbf{h}^T_t, \mathbf{c}_t ) )
\end{equation}
where~$g(\cdot)$ is a one-hidden-layer neural network with the
following parametrization:
\begin{equation}
g( \mathbf{h}^T_t, \mathbf{c}_t ) = \mathbf{W}_o \tanh (\mathbf{U}_h \mathbf{h}^T_t + \mathbf{W}_h \mathbf{c}_t )
\end{equation}
where $\mathbf{W}_o \in \mathbb{R}^{|V| \times d}$, $\mathbf{U}_h \in
\mathbb{R}^{d \times d}$, and $\mathbf{W}_h \in \mathbb{R}^{d \times
	d}$; $|V|$ is the output vocabulary size and $d$ the hidden unit
size.  $\mathbf{h}^T_t$ is the hidden state of the decoder LSTM which
summarizes~$y_{1:t}$, i.e.,~what has been generated so far:
\begin{equation}
\mathbf{h}^T_t = \text{LSTM}(y_t, \mathbf{h}^T_{t-1})
\end{equation}
The dynamic context vector $\mathbf{c}_t$ is the weighted sum of the
hidden states of the source sentence:
\begin{equation}
\mathbf{c}_t = \sum_{i=1}^{|X|} \alpha_{ti} \mathbf{h}^S_i
\end{equation}
whose weights~$\alpha_{ti}$ are determined by an \emph{attention}
mechanism:
\begin{equation}
\label{eq:attention}
\alpha_{ti} = \frac{\exp( \mathbf{h}^T_t \boldsymbol{\cdot}  \mathbf{h}^S_i )}{\sum_i \exp( \mathbf{h}^T_t \boldsymbol{\cdot}  \mathbf{h}^S_i )}
\end{equation}
where $\boldsymbol{\cdot}$ is the dot product between two vectors. We
use the dot product here mainly for efficiency reasons; alternative
ways to compute attention scores have been proposed in the literature
and we refer the interested reader to
\newcite{luong-etal:2015:EMNLP}. The model sketched above is usually
trained by minimizing the negative log-likelihood of the training
source-target pairs.

\section{Reinforcement Learning for Sentence Simplification}
\label{sec:rf}
In this section we present {\sc Dress}, our {\bf D}eep {\bf
  RE}inforcement {\bf S}entence {\bf S}implification model.  Despite
successful application in numerous sequence transduction tasks
\cite{jean2015montreal,chopra-auli-rush:2016:N16-1,xu2015show}, a
vanilla encoder-decoder model is not ideal for sentence
simplification. Although a number of rewrite operations
(e.g.,~copying, deletion, substitution, word reordering) can be used
to simplify text, copying is by far the most common. We empirically
found that~73\% of the target words are copied from the source in the
Newsela dataset. This number further increases to~83\% when
considering Wikipedia-based datasets (we provide details on these
datasets in Section~\ref{sec:dataset}). As a result, a generic
encoder-decoder model learns to copy all too well at the expense of
other rewrite operations, often parroting back the source or making
only a few trivial changes.

To encourage a wider variety of rewrite operations while remaining
fluent and faithful to the meaning of the source, we employ a
reinforcement learning framework (see
Figure~\ref{fig:model-overview}).  We view the encoder-decoder model
as an agent which first reads the source sentence~$X$; then at each
step, it takes an action $\hat{y}_t \in V$ (where $V$ is the output
vocabulary) according to a policy
$P_{RL}(\hat{y_t}|\hat{y}_{1:t-1}, X)$ (see
Equation~(\ref{eq:policy})). The agent continues to take actions until
it produces an \textbf{E}nd {\bf O}f {\bf S}entence (EOS) token
yielding the action sequence
$\hat{Y}=(\hat{y}_1, \hat{y}_2, \dots, \hat{y}_{|\hat{Y}|})$, which is
also the simplified output of our model. A reward~$r$ is then received
and the REINFORCE algorithm \cite{williams1992simple} is used to
update the agent.  In the following, we first introduce our reward and
then present the details of the REINFORCE algorithm.

\subsection{Reward}
\label{sec:reward}

The reward~$r(\hat{Y})$ for system output~$\hat{Y}$ is the weighted sum of
the three components aimed at capturing key aspects of the target
output, namely simplicity, relevance, and fluency:
\begin{equation}
r(\hat{Y}) = \lambda^S \, r^S + \lambda^R \, r^R +  \lambda^F \, r^F
\end{equation}
where $\lambda^S, \lambda^R, \lambda^F \in [0, 1]$; $r(\hat{Y})$ is a
shorthand for $r(X, Y, \hat{Y})$ where $X$ is the source, $Y$ the
reference (or target), and $\hat{Y}$ the system output. $r^S$, $r^R$,
and $r^F$ are shorthands for simplicity $r^S(X, Y, \hat{Y})$,
relevance $r^R(X,\hat{Y})$, and fluency~$r^F(\hat{Y})$. We provide
details for each reward summand below.

\paragraph{Simplicity} To encourage the model to apply a wide range of
simplification operations, we use SARI \cite{Xu_TACL16}, a recently
proposed metric which compares \textbf{S}ystem output \textbf{A}gainst
\textbf{R}eferences and against the \textbf{I}nput sentence. SARI is
the arithmetic average of n-gram precision and recall of three rewrite
operations: addition, copying, and deletion. It rewards addition
operations where system output was not in the input but occurred in
the references. Analogously, it rewards words retained/deleted in both
the system output and the references. In experimental evaluation
\newcite{Xu_TACL16} demonstrate that SARI correlates well with human
judgments of simplicity, whilst correctly rewarding systems that both
make changes and simplify the input.

One caveat with using SARI as a reward is the fact that it relies on
the availability of multiple references which are rare for sentence
simplification. \newcite{Xu_TACL16} provide eight references for
2,350~sentences, but these are primarily for system tuning and
evaluation rather than training.  The majority of existing
simplification datasets (see Section~\ref{sec:dataset} for details)
have a single reference for each source sentence. Moreover, they are
unavoidably noisy as they are mostly constructed automatically,
e.g.,~by aligning sentences from the ordinary and simple English
Wikipedias. When relying solely on a single reference, SARI will try
to reward accidental n-grams that should never have occurred in it. To
countenance the effect of noise, we apply
$\textsc{Sari}(X, \hat{Y}, Y)$ in the expected direction, with~$X$ as
the source, $\hat{Y}$ the system output, and $Y$ the reference as well
as in the reverse direction with $Y$ as the system output and
$\hat{Y}$ as the reference.  Assuming our system can produce
reasonably good simplifications, by swapping the output and the
reference, reverse SARI can be used to estimate how good a reference
is with respect to the system output.  Our first reward is therefore
the weighted sum of SARI and reverse SARI:
\begin{equation}
\label{eq:auto}
r^S \hspace*{-1ex}=\hspace*{-.3ex}  \beta \,  \textsc{\small Sari}(X, \hat{Y}, Y) 
+ (1 - \beta) \, \textsc{\small Sari}(X, Y, \hat{Y})
\end{equation}

\paragraph{Relevance} While the simplicity-based reward~$r^S$ tries to
encourage the model to make changes, the relevance reward $r^R$
ensures that the generated sentences preserve the meaning of the
source. We use an LSTM sentence encoder to convert the source~$X$ and
the predicted target~$\hat{Y}$ into two vectors~$\mathbf{q}_X$
and~$\mathbf{q}_{\hat{Y}}$. The relevance reward $r^{R}$ is simply the
cosine similarity between these two vectors:
\begin{equation}
\label{eq:relevance}
r^{R} = \cos( \mathbf{q}_{X}, \mathbf{q}_{\hat{Y}} ) = \frac{  \mathbf{q}_{X} \boldsymbol{\cdot}  
	\mathbf{q}_{\hat{Y}} }{ || \mathbf{q}_{X} || \, || \mathbf{q}_{\hat{Y}} || }  
\end{equation}

We use a sequence auto-encoder (SAE; \citealt{dai2015semi}) to train
the LSTM sentence encoder on both the complex and simple
sentences. Specifically, the SAE uses sentence $X=(x_1,\dots,x_{|X|})$
to infer itself via an encoder-decoder model (without an attention
mechanism).  Firstly, an encoder LSTM converts~$X$ into a sequence of
hidden states $(\mathbf{h}_1,\dots, \mathbf{h}_{|X|})$. Then, we
use~$\mathbf{h}_{|X|}$ to initialize the hidden state of the decoder
LSTM and recover/generate~$X$ one word at a time. 

\paragraph{Fluency} \newcite{Xu_TACL16} observe that SARI correlates
less with fluency compared to other metrics such as BLEU
\cite{papineni:2002}.  The fluency reward~$r^F$ models the
well-formedness of the generated sentences explicitly. It is the
normalized sentence probability assigned by an LSTM language model
trained on simple sentences:
\begin{equation}
r^F = \exp \left( \frac{1}{|\hat{Y}|} \sum_{i=1}^{|\hat{Y}|} \log P_{LM}(\hat{y}_i|\hat{y}_{0:i-1}) \right)
\end{equation}
We take the exponential of~$\hat{Y}$'s perplexity to ensure that
\mbox{$r^F \in [0, 1]$} as is the case with~$r^S$ and~$r^R$.

\subsection{The REINFORCE Algorithm}
The goal of the REINFORCE algorithm is to find an agent that maximizes
the expected reward. The training loss for one sequence is its 
negative expected reward:
\begin{equation*}
\mathcal{L}(\theta) = - \mathbb{E}_{ (\hat{y}_1,\dots,\hat{y}_{|\hat{Y}|}) \sim P_{RL}(\cdot|X)} [ r( \hat{y}_1,{\small \dots},\hat{y}_{|\hat{Y}|} ) ]
\end{equation*}
where $P_{RL}$ is our policy, i.e.,~the distribution produced by the
encoder-decoder model (see Equation\eqref{eq:policy}) and
$r(\cdot)$~is the reward function of an action sequence
$\hat{Y} = ( \hat{y}_1,\dots,\hat{y}_{|\hat{Y}|} )$, i.e.,~a generated
simplification. Unfortunately, computing the expectation term is
prohibitive, since there is an infinite number of possible action
sequences. In practice, we approximate this expectation with a single
sample from the distribution of $P_{LR}(\cdot|X)$. We refer to
\newcite{williams1992simple} for the full derivation of the
gradients. The gradient of $\mathcal{L}(\theta)$ is:
\[
\begin{array}{ll}
\nabla\mathcal{L}(\theta)\approx \\
\sum_{t=1}^{|\hat{Y}|}\nabla \log P_{RL}(\hat{y}_t|\hat{y}_{1:t-1}, X) [r(\hat{y}_{1:|\hat{Y}|}) - b_t ]
\end{array}
\]
To reduce the variance of gradients, we also introduce a baseline
linear regression model $b_t$ to estimate the expected future reward
at time $t$ \cite{ranzato2016sequence}. $b_t$~takes the concatenation
of~$\mathbf{h}_t^T$ and $\mathbf{c}_t$ as input and outputs a real
value as the expected reward. The parameters of the regressor are
trained by minimizing mean squared error. We do not back-propagate
this error to~$\mathbf{h}_t^T$ or $\mathbf{c}_t$ during training
\cite{ranzato2016sequence}.

\subsection{Learning}
\label{sec:learning}
Presented in its original form, the REINFORCE algorithm starts
learning with a random policy. This assumption can make model training
challenging for generation tasks like ours with large vocabularies
(i.e.,~action spaces). We address this issue by pre-training our agent
(i.e.,~the encoder-decoder model) with a negative log-likelihood
objective (see Section \ref{sec:encdeca}), making sure it can produce
reasonable simplifications, thereby starting off with a policy which
is better than random. We follow prior work \cite{ranzato2016sequence}
in adopting a curriculum learning strategy. In the beginning of
training, we give little freedom to our agent allowing it to predict
the last few words for each target sentence. For every target sequence, we
use negative log-likelihood to train the first $L$ (initially,~$L=24$)
tokens and apply the reinforcement learning algorithm to the $(L+1)$th
tokens onwards. Every two epochs, we set \mbox{$L=L-3$} and the
training terminates when $L$ is 0.


\section{Lexical Simplification}
\label{sec:ltm}
Lexical substitution, the replacement of complex words with simpler
alternatives, is an integral part of sentence simplification
\cite{specia-jauhar-mihalcea:2012:STARSEM-SEMEVAL}.
The model presented so far learns lexical substitution and other
rewrite operations \emph{jointly}. In some cases, words are predicted because
they seem natural in the their context, but are poor substitutes for
the content of the complex sentence. To countenance this, we learn
lexical simplifications \emph{explicitly} and integrate them with our
reinforcement learning-based model.




We use an {\it pre-trained} encoder-decoder model (which is trained on a parallel corpus of
complex and simple sentences) to obtain probabilistic word alignments,
aka attention scores (see $\alpha_t$ in
Equation~(\ref{eq:attention})).  Let $X = (x_1, x_2, \dots, x_{|X|})$
denote a source sentence and $Y = (y_1, y_2, \dots, y_{|Y|})$ a target
sentence. We convert~$X$ into~$|X|$ hidden states
$(\mathbf{v}_1,\mathbf{v}_2, \dots, \mathbf{v}_{|X|})$ with an
LSTM. Note that \mbox{$\mathbf{v}_t \in \mathbb{R}^{d \times 1}$}
corresponds to the context dependent representation of~$x_t$. Let
$\alpha_t$ denote the alignment scores
$\alpha_{t1}, \alpha_{t2}, \dots, \alpha_{t|X|}$. The lexical
simplification probability of~$y_t$ given the source sentence and the
alignment scores is:
\begin{equation}
P_{LS}(y_t | X, \alpha_t) = \text{softmax}(\mathbf{W}_l \, \mathbf{s}_t) 
\end{equation}
where $\mathbf{W}_l \in \mathbb{R}^{|V| \times d}$ and $\mathbf{s}_t$
represents the source:
\begin{equation}
\mathbf{s}_t = \sum_{i=1}^{|X|} \alpha_{ti} \mathbf{v}_i
\end{equation}

The lexical simplification model on its own encourages lexical
substitutions, without taking into account what has been generated so
far (i.e.,~$y_{1:t-1}$) and as a result fluency could be
compromised. A straightforward solution is to integrate lexical
simplification with our reinforcement learning trained model
(Section~\ref{sec:rf}) using linear interpolation, where
$\eta \in [0, 1]$:
\begin{equation}
\begin{split}
\hspace*{-3ex}P(y_t|y_{1:t-1}, X) = \, & (1-\eta) \, P_{RL}(y_t|y_{1:t-1},X) \\
\hspace*{-8ex}		& + \eta \, P_{LS}(y_t|X, \alpha_t)
\end{split}\hspace*{-.4cm}
\end{equation}

\section{Experimental Setup}
\label{sec:experimental-setup}

In this section we present our experimental setup for assessing the
performance of the simplification model described above. We give
details on our datasets, model training,  evaluation
protocol, and the systems used for comparison.

\paragraph{Datasets}
\label{sec:dataset}

We conducted experiments on three simplification datasets.
\textit{WikiSmall} \cite{zhu2010monolingual} is a parallel corpus
which has been extensively used as a benchmark for evaluating text
simplification systems
\cite{wubben2012sentence,woodsend-lapata:2011:EMNLP,narayan-gardent:2014,zhu2010monolingual}. It
contains automatically aligned complex and simple sentences from the
ordinary and simple English Wikipedias. The test set consists of
100~complex-simple sentence pairs. The training set contains 89,042
sentence pairs (after removing duplicates and test sentences).  We
randomly sampled 205~pairs for development and used the
remaining sentences for training.  

We also constructed \textit{WikiLarge}, a larger Wikipedia corpus by
combining previously created simplification corpora. Specifically, we
aggregated the aligned sentence pairs in
\newcite{kauchak:2013:ACL2013}, the aligned and revision sentence
pairs in \newcite{woodsend-lapata:2011:EMNLP}, and Zhu's
(\citeyear{zhu2010monolingual}) WikiSmall dataset described above. We
used the development and test sets created in
\newcite{Xu_TACL16}. These are complex sentences taken from WikiSmall
paired with simplifications provided by Amazon Mechanical Turk
workers. The dataset contains 8 (reference) simplifications for 2,359
sentences partitioned into~2,000 for development and 359~for
testing. After removing duplicates and sentences in development and
test sets, the resulting training set contains 296,402 sentence pairs.



Our third dataset is \textit{Newsela}, a corpus collated by
\newcite{Xu-EtAl:2015:TACL} who argue that Wikipedia-based resources
are suboptimal due to the automatic sentence alignment which
unavoidably introduces errors, and their uniform writing style which
leads to systems that generalize poorly.
Newsela\footnote{\url{https://newsela.com}} consists of 1,130 news articles,
each re-written four times by professional editors for children at
different grade levels (0 is the most complex level and 4~is
simplest). \newcite{Xu-EtAl:2015:TACL} provide multiple aligned
complex-simple pairs within each article.  We removed sentence pairs
corresponding to levels~\mbox{0--1}, \mbox{1--2}, and \mbox{2--3},
since they were too similar to each other. The first 1,070 documents
were used for training (94,208 sentence pairs), the next 30 documents
for development (1,129 sentence pairs) and the last~30 documents for
testing (1,076 sentence pairs).\footnote{If a sentence has multiple
	references in the development or test set, we use the reference with
	highest simplicity level.}  We are not aware of any published
results on this dataset.


\paragraph{Training Details}
We trained our models on an Nvidia GPU card. We used the
same hyper-parameters across datasets.  We first trained an
encoder-decoder model, and then performed reinforcement learning
training (Section~\ref{sec:rf}), and trained the lexical
simplification model (Section~\ref{sec:ltm}).  Encoder-decoder
parameters were uniformly initialized to~$[-0.1, 0.1]$.  We used Adam
\cite{kingma:2014} to optimize the model with learning rate~0.001; the
first momentum coefficient was set to~0.9 and the second momentum
coefficient to~0.999. The gradient was rescaled when the norm
exceeded~5 \cite{pascanu:2013}. Both encoder and decoder LSTMs have
two layers with 256 hidden neurons in each layer. We regularized all
LSTMs with a dropout rate of~0.2
\cite{zaremba:2014}. We initialized the encoder and
decoder word embedding matrices with  300~dimensional Glove
vectors \cite{pennington:2014}.

During reinforcement training, we used plain stochastic gradient
descent with a learning rate of~0.01. We set $\beta = 0.1$,
$\lambda_S = 1$, $\lambda_R = 0.25$ and
$\lambda_F = 0.5$.\footnote{Weights were tuned on the development set
	of the Newsela dataset and kept fixed for the other two datasets.}
Training details for the lexical simplification model are identical to
the encoder-decoder model except that word embedding matrices were
randomly initialized. The weight of the lexical simplification model
was set to~$\eta=0.1$.

To reduce vocabulary size, named entities were tagged with the
Stanford CoreNLP \cite{manning-EtAl:2014:P14-5} and anonymized with a
$\text{NE}@N$ token, where
$\text{NE} \in \{\text{PER}, \text{LOC}, \text{ORG}, \text{MISC} \}$
and $N$ indicates $\text{NE}@N$ is the $N$-th distinct NE typed
entity. For example, ``John and Bob are \dots'' becomes ``PER@1 and
PER@2 are \dots''. At test time, we de-anonymize~$\text{NE}@N$ tokens
in the output by looking them up in their source sentences. Note that
the de-anonymization may fail, but the chance is small (around 2\%~of
the time on the Newsela development set). We replaced words occurring
three times or less in the training set with UNK. At test time, when
our models predict UNK, we adopt the UNK replacement method proposed
in \citet{jean2015montreal}.

\paragraph{Evaluation}
Following previous work \cite{woodsend-lapata:2011:EMNLP,Xu_TACL16} we
evaluated system output automatically adopting metrics widely used in
the simplification literature. Specifically, we used
BLEU\footnote{With the default {\tt mtevalv13a.pl} settings.}
\cite{papineni:2002} to assess the degree to which generated
simplifications differed from gold standard references and the
Flesch-Kincaid Grade Level index (FKGL; \citealt{kincaid:1975}) to
measure the readability of the output (lower FKGL\footnote{FKGL implementation at {\small \url{http://goo.gl/OHP7k3}}.} implies
simpler output). In addition, we used SARI \cite{Xu_TACL16}, which
evaluates the quality of the output by comparing it against the source
and reference simplifications.\footnote{We used he implementation of
  SARI in \newcite{Xu_TACL16}.}  BLEU, FKGL, and SARI are all measured
at corpus-level.
We also evaluated system output by eliciting human judgments via
Amazon's Mechanical Turk. Specifically (self-reported) native English
speakers were asked to rate simplifications on three dimensions:
\emph{Fluency} (is the output grammatical and well formed?),
\emph{Adequacy} (to what extent is the meaning expressed in the
original sentence preserved in the output?) and \emph{Simplicity} (is
the output simpler than the original sentence?). All ratings were
obtained using a five point Likert scale.

\paragraph{Comparison Systems}
\label{sec:model:comp}
We compared our model against several systems previously proposed in
the literature. These include PBMT-R, a monolingual phrase-based
machine translation system with a reranking post-processing
step\footnote{We made a good-faith effort to re-implement their system
  following closely the details in \newcite{wubben2012sentence}.}
\cite{wubben2012sentence} and Hybrid, a model which first performs
sentence splitting and deletion operations over discourse
representation structures and then further simplifies sentences with
{PBMT-R} \cite{narayan-gardent:2014}. Hybrid\footnote{We are grateful
  to Shashi Narayan for running his system on our three datasets.} is
state of the art on the WikiSmall dataset. Comparisons with
\mbox{SBMT-SARI}, a syntax-based translation model trained on PPDB
\cite{ganitkevitch2013ppdb} and tuned with SARI \cite{Xu_TACL16}, are
problematic due to the size of PPDB which is considerably larger than
any of the datasets used in this work (it contains 106~million
sentence pairs with 2~billion words). Nevertheless, we
compare\footnote{The output of SBMT-SARI is publicly available.}
against SBMT-SARI, but only models trained on Wikilarge, our largest
dataset.



\section{Results}
\label{sec:results}

\begin{table}[t]
	\centering
	\setlength{\belowcaptionskip}{-.5cm}
	\small
	\begin{tabular}{| l | c c c | c c c |}
		\hline
		\multicolumn{1}{|l|}{Newsela} & BLEU & FKGL & SARI \\
		\hline
		\hline
		PBMT-R  & 18.19 & 7.59 & 15.77 \\
		Hybrid & 14.46 & {\bf 4.01} & {\bf 30.00} \\
		EncDecA & 21.70 & 5.11 & 24.12 \\
		{\sc Dress}  & 23.21 & 4.13 & {27.37} \\
		{\sc Dress-Ls}  & \bf{24.30} & 4.21 & 26.63 \\
		\hline
		\multicolumn{4}{c}{} \\ \hline
		WikiSmall &  BLEU & FKGL & SARI \\ \hline \hline
		PBMT-R &  46.31 & 11.42 & 15.97 \\
		Hybrid &  \bf{53.94} & \hspace*{.8ex}9.20 & \bf{30.46} \\
		EncDecA &  {47.93} & 11.35 & 13.61 \\
		{\sc Dress} &  34.53 & \hspace*{.8ex}\bf{7.48} & 27.48 \\
		{\sc Dress-Ls} & 36.32 & \hspace*{.8ex}7.55 & 27.24 \\ \hline
		\multicolumn{4}{c}{} \\ \hline
		WikiLarge &  BLEU & FKGL & SARI \\ \hline \hline
		PBMT-R & 81.11 & 8.33 & 38.56 \\
		Hybrid & 48.97 & {\bf 4.56} & 31.40 \\
		SBMT-SARI & 73.08 & 7.29 & {\bf 39.96} \\
		EncDecA & {\bf 88.85} & 8.41 & 35.66 \\
		{\sc Dress} & 77.18  & 6.58 & 37.08 \\
		{\sc Dress-Ls} & 80.12 & 6.62 & 37.27 \\
		\hline
	\end{tabular}
	\caption{Automatic evaluation on Newsela, WikiSmall, and WikiLarge
		test sets.}
	\label{tbl:auto-newsela}
\end{table}

\begin{table}[t]
	\centering
	\setlength{\belowcaptionskip}{-.5cm}
	\small
	\begin{tabular}{|@{~} l@{~} |@{~} c@{~} c@{~}  c@{~}  c@{~} |}
		\hline
		Newsela & Fluency & Adequacy & Simplicity & All \\
		\hline
		\hline
		PBMT-R & 3.56 & \hspace*{2ex}\bf{3.58}$^{**}$& \hspace*{2ex}2.09$^{**}$ & \hspace*{2ex}3.08$^{**}$ \\
		Hybrid & \hspace*{2ex}2.70$^{**}$ & \hspace*{2ex}2.51$^{**}$ & 2.99 & \hspace*{2ex}2.73$^{**}$ \\
		EncDecA & 3.63 & 2.99 & \hspace*{2ex}2.56$^{**}$ & \hspace*{2ex}3.06$^{**}$ \\
		{\sc Dress} & 3.65 & 2.94 & \bf{3.10} & 3.23 \\
		{\sc Dress-Ls} & \bf{3.71} & 3.07 & 3.04 & \bf{3.28} \\
		Reference & 3.90 & \hspace*{2ex}2.81$^{**}$ & \hspace*{2ex}3.42$^{**}$ & 3.38 \\ \hline
		\multicolumn{4}{c}{} \\\hline
		WikiSmall & Fluency & Adequacy & Simplicity & All \\
		\hline
		\hline
		PBMT-R & 3.91 & \hspace*{2ex}{\bf 3.74}$^{**}$ & \hspace{2ex}2.80$^{**}$ & \hspace{1ex}3.48$^{*}$ \\
		Hybrid & \hspace*{2ex}3.26$^{**}$ & 3.42 & \hspace{2ex}2.82$^{**}$ & \hspace{2ex}3.17$^{**}$ \\
		{\sc Dress-Ls} & \bf{3.92} & 3.36 & \bf{3.55} & \bf{3.61} \\
		Reference & \hspace{1ex}3.74$^{*}$ & 3.34 & \hspace{2ex}3.13$^{**}$ & \hspace{2ex}3.41$^{**}$ \\\hline
		\multicolumn{4}{c}{} \\\hline
		
		WikiLarge & Fluency & Adequacy & Simplicity & All \\\hline \hline
		PBMT-R & 3.68 & \hspace{1ex}{\bf 3.63}$^{*}$ & \hspace{2ex}2.70$^{**}$ & \hspace{1ex}3.34$^{*}$ \\
		Hybrid & \hspace{2ex}2.60$^{**}$ & \hspace{2ex}2.42$^{**}$ & {\bf 3.52} & \hspace{2ex}2.85$^{**}$ \\
		SBMT-SARI & \hspace{2ex}3.34$^{**}$ & \hspace{1ex}3.51$^{*}$ & \hspace{2ex}2.77$^{**}$ & \hspace{2ex}3.21$^{**}$ \\
		{\sc Dress-Ls} & {\bf 3.70} & 3.28 & {3.42} & {\bf 3.46}
		\\
		Reference & 3.79 & \hspace{2ex}3.72$^{**}$ & \hspace{2ex}2.86$^{**}$ & 3.46 \\
		\hline
	\end{tabular}
	\caption{Mean ratings elicited by humans on Newsela, WikiSmall,
		and WkiLarge test sets. Ratings
		significantly different from {\sc Dress-Ls} are marked
		with * (\mbox{$p<0.05$}) and ** (\mbox{$p<0.01$}). Significance tests were performed using a student
		\mbox{$t$-test}.} 
	\label{tbl:human-newsela}
\end{table}

Since Newsela contains high quality simplifications created by
professional editors, we performed the bulk of our experiments on this
dataset. Specifically, we set out to answer two questions: (a)~which
neural model performs best and (b)~how do neural models which are
resource lean and do not have access to linguistic annotations fare
against more traditional systems. We therefore compared the basic
attention-based encoder-decoder model (EncDecA), with the deep
reinforcement learning model ({\sc Dress}; Section~\ref{sec:rf}), and a
linear combination of {\sc Dress} and the lexical simplification model
({\sc Dress-Ls}; Section~\ref{sec:ltm}).  Neural models were further
compared against two strong baselines, PBMT-R and Hybrid.
Table~\ref{tbl:example} shows example output of all models on the
Newsela dataset.

\begin{table*}[t]
	\centering
	\begin{tabular}{| l  p{13cm}@{~} |}
		\hline
		Complex & There's just one major hitch: the primary
		purpose of education is to develop citizens with a
		wide variety of skills. \\ 	
		Reference & The purpose of education is to {develop} a wide range of skills. \\ 
		PBMT-R & It's just one major hitch: the purpose of
		education is to {\bf make people} with a wide variety
		of skills. \\  
		Hybrid & one hitch the purpose is to develop citizens. \\
		EncDecA & The {\bf key} of education is to develop {\bf people} with a wide
		variety of skills. \\ 
		{\sc Dress} & There's just one major hitch: the {\bf main goal} of
		education is to develop {\bf people} with {\bf lots of} skills. \\
		{\sc Dress-Ls} & There's just one major hitch: the {\bf main goal} of education is to develop citizens with {\bf lots of} skills. \\\hline
		\hline
		Complex & {``They were so burdened by the past they couldn't think about the future,'' said Barnet, 62, who was president of Columbia Records, the No.1 record label in the United States, before joining Capitol.} \\ 
		Reference & Capitol was stuck in the past. It could not think about the future, Barnett said. \\ 
		PBMT-R & ``They were so {\bf affected} by the past they couldn't think about the future,'' said Barnett, 62, was president of Columbia Records, before joining Capitol {\bf building}. \\
		Hybrid &  `They were so burdened by the past they couldn't think about the future,'' said Barnett, 62, who was Columbia Records, president of the No.1 record label in the united states, before joining Capitol. \\
		EncDecA &  ``They were so burdened by the past they couldn't think about the future,'' said Barnett, who was president of Columbia Records, the No.1 record labels in the United States. \\
		{\sc Dress} & ``They were so {\bf sicker} by the past they couldn't think about the future,'' said Barnett, who was president of Columbia Records. \\
		{\sc Dress-Ls} &  ``They were so burdened by the past they couldn't think about the future,'' said Barnett, who was president of Columbia Records. \\
		\hline
	\end{tabular}
	\caption{System output for two sentences (Newsela
          development set). Substitutions are shown in bold.}
	\label{tbl:example}
\end{table*}

The top block in Table~\ref{tbl:auto-newsela} summarizes the results
of our automatic evaluation. As can be seen, all neural models obtain
higher BLEU, lower FKGL and higher SARI compared to PBMT-R. Hybrid has
the lowest FKGL and highest SARI.  Compared to EncDecA, {\sc Dress} scores
lower on FKGL and higher on SARI, which indicates that the model has
indeed learned to optimize the reward function which includes SARI.
Integrating lexical simplification ({\sc Dress-Ls}) yields better BLEU, but
slightly worse FKGL and SARI. 

The results of our human evaluation are presented in the top block of
Table~\ref{tbl:human-newsela}. We elicited judgments for 100~randomly
sampled test sentences. Aside from comparing system output (PBMT-R,
Hybrid, EncDecA, {\sc Dress}, and \mbox{{\sc Dress-Ls}}), we also elicited ratings
for the gold standard Reference as an upper bound.  We report results
for Fluency, Adequacy, and Simplicity individually and in combination
(All is the average rating of the three dimensions). 
As can be seen, {\sc Dress} and \mbox{{\sc Dress-Ls}} outperform \mbox{PBMT-R} and Hybrid on Fluency, Simplicity, and overall. The fact that neural
models (EncDecA, {\sc Dress} and \mbox{{\sc Dress-Ls}}) fare well on Fluency, is perhaps not surprising given the
recent success of LSTMs in language modeling and neural machine
translation \cite{zaremba:2014,jean2015montreal}. 

Neural models obtain worse ratings on Adequacy but are closest to the
human references on this dimension.  {\sc Dress-Ls} (and {\sc Dress}) are
significantly better ($p<0.01$) on Simplicity than EncDecA, PBMT-R,
and Hybrid which indicates that our reinforcement learning based model
is effective at creating simpler output. Combined ratings (All) for
{\sc Dress-Ls} are significantly different compared to the other models but
not to {\sc Dress} and the Reference. Nevertheless, integration of the
lexical simplification model boosts performance as ratings increase almost
across the board (Simplicity is slightly worse). Returning to our original questions, we find that
neural models are more fluent than comparison systems, while
performing non-trivial rewrite operations (see the SARI scores in
Table~\ref{tbl:auto-newsela}) which yield simpler output (see the
Simplicity column in Table~\ref{tbl:human-newsela}). Based on our
judgment elicitation study, neural models trained with reinforcement
learning perform best, with \mbox{{\sc Dress-Ls}} having a slight advantage.

%



We further analyzed model performance by computing various statistics
on the simplified output. We measured average sentence length and the
degree to which \textsc{Dress} and comparison systems perform
rewriting operations. We approximated the latter with Translation
Error Rate (TER; \citealt{snover2006study}), a measure commonly used
to automatically evaluate the quality of machine translation
output. We used TER to compute the (average) number of edits required
to change an original complex sentence to simpler output. We also
report the number of edits by type, i.e., the number of insertions,
substitutions, deletions, and shifts needed (on average) to convert
complex to simple sentences.  

As shown in Table~\ref{tab:lenter}, Hybrid obtains the highest TER,
followed by our models ({\sc Dress} and {\sc Dress-Ls}), which
indicates that they actively perform rewriting. Perhaps Hybrid is too
aggressive when simplifying a sentence, it obtains low Fluency and
Adequacy scores in human evaluation
(Table~\ref{tbl:human-newsela}). There is a strong correlation between
sentence length and number of deletion operations (i.e., more
deleteions lead to shorter sentences) and PBMT-R performs very few
deletions. Overall, reinforcement learning encourages deletion (see
{\sc Dress} and {\sc Dress-Ls}), while performing a reasonable amount
of additional operations (e.g.,~substitutions and shifts) compared to
EncDecA and PBMT-R.

\begin{table}[t]
	\small
	\centering
	\begin{tabular}{|l | c  c  c  c  c  c|}
		\hline
		Models & Len & TER & Ins & Del & Sub & Shft \\
		\hline
		\hline
		PBMT-R & 23.1 & 0.13 & 0.68 & 0.68 & 1.50 & 0.09 \\
		Hybrid & 12.4 & 0.90 & 0.01 & 10.19 & 0.12 & 0.41 \\
		EncDecA & 17.0 & 0.36 & 0.13 & 5.96 & 1.69 & 0.09 \\
		{\sc Dress} & 14.2 &  0.46 & 0.07 & 8.53 & 1.37 & 0.11 \\
		{\sc Dress-Ls} & 14.4 &  0.44 & 0.07 & 8.38 & 1.11 & 0.09 \\
		Reference & 12.7 & 0.67 & 0.40 & 10.26 & 3.44 & 0.73 \\
		\hline
	\end{tabular}
	\caption{Output length (average number of
          tokens), TER scores and number of edits by type
          (Ins\emph{ertions}, Del\emph{etions}, Sub\emph{stitutions},
          Sh\emph{i}ft\emph{s}) on the Newsela test set. Higher TER 
          means that more rewriting operations are performed.}
	\label{tab:lenter}
\end{table}

The middle blocks in Tables~\ref{tbl:auto-newsela}
and~\ref{tbl:human-newsela} report results on the WikiSmall
dataset. FKGL and SARI follow a similar pattern as on Newsela. BLEU
scores for PBMT-R, Hybrid, and EncDecA are much higher compared to
{\sc Dress} and \mbox{{\sc Dress-Ls}}. Hybrid obtains best BLEU and
SARI scores, while {\sc Dress} and \mbox{{\sc Dress-Ls}} do very well
on FKGL.  In human evaluation, we elicited judgments on the entire
WikiSmall test set (100 sentences). We compared \mbox{{\sc Dress-Ls}},
with \mbox{PBMT-R}, Hybrid, and gold standard Reference
simplifications. As human experiments are time consuming and
expensive, we did not include other neural models besides {\sc
  Dress-Ls} based on our Newsela study which showed that EncDecA is
inferior to variants trained with reinforcement learning and that {\sc
  Dress-Ls} is the better performing model (however, we do compare
\emph{all} models in Table~\ref{tbl:auto-newsela}). \mbox{{\sc
    Dress-Ls}} is significantly better on Simplicity than
\mbox{PBMT-R}, Hybrid, and the Reference.  It performs on par with
PBMT-R on Fluency and worse on Adequacy (but still closer to the human
Reference than \mbox{PBMT-R} or Hybrid). When combining all ratings
(All in Table~\ref{tbl:human-newsela}), \mbox{{\sc Dress-Ls}} is
significantly better than PBMT-R, Hybrid, and the \mbox{Reference}.

The bottom blocks in Tables~\ref{tbl:auto-newsela}
and~\ref{tbl:human-newsela} report results on Wikilarge. We compared
our models with PBMT-R, Hybrid, and \mbox{SBMT-SARI}
\cite{Xu_TACL16}. The FKGL follows a similar pattern as in the
previous datasets. PBMT-R and our models are best in terms of BLEU
while \mbox{SBMT-SARI} outperforms all other systems on
SARI.\footnote{BLEU and SARI scores reported in \newcite{Xu_TACL16}
  are 72.36 and 37.91, and measured at sentence-level.} Because there
are 8 references for each complex sentence in the test set, BLEU
scores are much higher compared to Newsela and WikiSmall. In human
evaluation, we again elicited judgments for 100~randomly sampled test
sentences. We randomly selected one of the 8 references as the
Reference upper bound. On Simplicity, \mbox{{\sc Dress-Ls}} is
significantly better than all comparison systems, except Hybrid. On
Adequacy, it is better than Hybrid but significantly worse than other
comparison systems. On Fluency, it is on par with PBMT-R\footnote{We
  used more data to train PBMT-R and maybe that is why PBMT-R performs
  better than \newcite{Xu_TACL16} reported.} but better than Hybrid
and SBMT-SARI. On All dimension \mbox{{\sc Dress-Ls}} significantly
outperforms all comparison systems.

\section{Conclusions}
We developed a reinforcement learning-based text simplification model,
which can jointly model simplicity, grammaticality, and semantic
fidelity to the input.  We also proposed a lexical simplification
component that further boosts performance. Overall, we find that
reinforcement learning offers a great means to inject prior knowledge
to the simplification task achieving good results across three
datasets. In the future, we would like to explicitly model sentence
splitting and simplify entire documents (rather than individual
sentences).  Beyond sentence simplification, the reinforcement
learning framework presented here is potentially applicable to
generation tasks such as sentence compression
\cite{chopra-auli-rush:2016:N16-1}, generation of programming code
\cite{ling-EtAl:2016:P16-1}, or poems
\cite{zhang-lapata:2014:EMNLP2014}.

\paragraph{Acknowledgments} We would like to thank Li Dong, Jianpeng
Cheng, Shashi Narayan and the EMNLP reviewers for their valuable
feedback. We are also grateful to Shashi Narayan for supplying us with
the output of his system and Wei Xu for her help with this work. The
authors acknowledge the support of the European Research Council
(award number 681760).

\bibliography{emnlp2017}

\begin{thebibliography}{49}
\expandafter\ifx\csname natexlab\endcsname\relax\def\natexlab#1{#1}\fi

\bibitem[{Bahdanau et~al.(2015)Bahdanau, Cho, and Bengio}]{bahdanau2014neural}
Dzmitry Bahdanau, Kyunghyun Cho, and Yoshua Bengio. 2015.
\newblock Neural machine translation by jointly learning to align and
  translate.
\newblock In \emph{Proceedings of ICLR}, San Diego, CA.

\bibitem[{{Beigman Klebanov} et~al.(2004){Beigman Klebanov}, Knight, and
  Marcu}]{Klebanov:ea:04}
Beata {Beigman Klebanov}, Kevin Knight, and Daniel Marcu. 2004.
\newblock Text simplification for information-seeking applications.
\newblock In \emph{Proceedings of ODBASE}, volume 3290 of \emph{Lecture Notes
  in Computer Science}, pages 735--747, Agia Napa, Cyprus. Springer.

\bibitem[{Carroll et~al.(1999)Carroll, Minnen, Pearce, Canning, Devlin, and
  Tait}]{Carroll:ea:99}
J.~Carroll, G.~Minnen, D.~Pearce, Y.~Canning, S.~Devlin, and J~Tait. 1999.
\newblock Simplifying text for language-impaired readers.
\newblock In \emph{Proceedings of the 9th EACL}, pages 269--270, Bergen,
  Norway.

\bibitem[{Chandrasekar et~al.(1996)Chandrasekar, Doran, and
  Srinivas}]{Chandrasekar:ea:96}
R.~Chandrasekar, C.~Doran, and B.~Srinivas. 1996.
\newblock Motivations and methods for text simplification.
\newblock In \emph{Proceedings of the 16th COLING}, pages 1041--1044,
  Copenhagen, Denmark.

\bibitem[{Chopra et~al.(2016)Chopra, Auli, and
  Rush}]{chopra-auli-rush:2016:N16-1}
Sumit Chopra, Michael Auli, and Alexander~M. Rush. 2016.
\newblock Abstractive sentence summarization with attentive recurrent neural
  networks.
\newblock In \emph{Proceedings of NAACL: HLT}, pages 93--98, San Diego, CA.

\bibitem[{Curran et~al.(2007)Curran, Clark, and
  Bos}]{curran-clark-bos:2007:PosterDemo}
James Curran, Stephen Clark, and Johan Bos. 2007.
\newblock Linguistically motivated large-scale nlp with c\&c and boxer.
\newblock In \emph{Proceedings of the 45th ACL Companion Volume Proceedings of
  the Demo and Poster Sessions}, pages 33--36, Prague, Czech Republic.

\bibitem[{Dai and Le(2015)}]{dai2015semi}
Andrew~M Dai and Quoc~V Le. 2015.
\newblock Semi-supervised sequence learning.
\newblock In \emph{Advances in Neural Information Processing Systems}, pages
  3079--3087.

\bibitem[{Devlin(1999)}]{Devlin:1999}
Siobhan Devlin. 1999.
\newblock \emph{Simplifying Natural Language for Aphasic Readers}.
\newblock Ph.D. thesis, University of Sunderland.

\bibitem[{Evans et~al.(2014)Evans, {a}san, and Dornescu}]{Evans:ea:2014}
Richard Evans, Constantin~Or\ {a}san, and Iustin Dornescu. 2014.
\newblock An evaluation of syntactic simplification rules for people with
  autism.
\newblock In \emph{Proceedings of the 3rd Workshop on Predicting and Improving
  Text Readability for Target Reader Populations (PITR)}, pages 131--140,
  Gothenburg, Sweden.

\bibitem[{Ganitkevitch et~al.(2013)Ganitkevitch, {Van Durme}, and
  Callison-Burch}]{ganitkevitch2013ppdb}
Juri Ganitkevitch, Benjamin {Van Durme}, and Chris Callison-Burch. 2013.
\newblock {PPDB}: The paraphrase database.
\newblock In \emph{Proceedings of NAACL-HLT}, pages 758--764, Atlanta, GA.

\bibitem[{Hochreiter and Schmidhuber(1997)}]{hochreiter1997long}
Sepp Hochreiter and J{\"u}rgen Schmidhuber. 1997.
\newblock Long short-term memory.
\newblock \emph{Neural computation}, 9(8):1735--1780.

\bibitem[{Inui et~al.(2003)Inui, Fujita, Takahashi, Iida, and
  Iwakura}]{inui-EtAl:2003:PARAPHRASE}
Kentaro Inui, Atsushi Fujita, Tetsuro Takahashi, Ryu Iida, and Tomoya Iwakura.
  2003.
\newblock Text simplification for reading assistance: A project note.
\newblock In \emph{Proceedings of the 2nd International Workshop on
  Paraphrasing}, pages 9--16, Sapporo, Japan.

\bibitem[{Jean et~al.(2015)Jean, Firat, Cho, Memisevic, and
  Bengio}]{jean2015montreal}
S{\'e}bastien Jean, Orhan Firat, Kyunghyun Cho, Roland Memisevic, and Yoshua
  Bengio. 2015.
\newblock Montreal neural machine translation systems for {WMT15}.
\newblock In \emph{Proceedings of the 10th Workshop on Statistical Machine
  Translation}, pages 134--140, Lisbon, Portugal.

\bibitem[{Kaji et~al.(2002)Kaji, Kawahara, Kurohashi, and
  Sato}]{kaji-EtAl:2002:ACL}
Nobuhiro Kaji, Daisuke Kawahara, Sadao Kurohashi, and Satoshi Sato. 2002.
\newblock Verb paraphrase based on case frame alignment.
\newblock In \emph{Proceedings of 40th ACL}, pages 215--222, Philadelphia, PA.

\bibitem[{Kauchak(2013)}]{kauchak:2013:ACL2013}
David Kauchak. 2013.
\newblock Improving text simplification language modeling using unsimplified
  text data.
\newblock In \emph{Proceedings of the 51st ACL}, pages 1537--1546, Sofia,
  Bulgaria.

\bibitem[{Kincaid et~al.(1975)Kincaid, Fishburne~Jr, Rogers, and
  Chissom}]{kincaid:1975}
J~Peter Kincaid, Robert~P Fishburne~Jr, Richard~L Rogers, and Brad~S Chissom.
  1975.
\newblock Derivation of new readability formulas (automated readability index,
  fog count and flesch reading ease formula) for navy enlisted personnel.
\newblock Technical report, DTIC Document.

\bibitem[{Kingma and Ba(2014)}]{kingma:2014}
Diederik Kingma and Jimmy Ba. 2014.
\newblock Adam: A method for stochastic optimization.
\newblock \emph{arXiv preprint arXiv:1412.6980}.

\bibitem[{Li et~al.(2016)Li, Monroe, Ritter, Jurafsky, Galley, and
  Gao}]{li-EtAl:2016:EMNLP}
Jiwei Li, Will Monroe, Alan Ritter, Dan Jurafsky, Michel Galley, and Jianfeng
  Gao. 2016.
\newblock Deep reinforcement learning for dialogue generation.
\newblock In \emph{Proceedings of the 2016 EMNLP}, pages 1192--1202, Austin,
  TX.

\bibitem[{Ling et~al.(2016)Ling, Blunsom, Grefenstette, Hermann,
  Ko\v{c}isk\'{y}, Wang, and Senior}]{ling-EtAl:2016:P16-1}
Wang Ling, Phil Blunsom, Edward Grefenstette, Karl~Moritz Hermann,
  Tom\'{a}\v{s} Ko\v{c}isk\'{y}, Fumin Wang, and Andrew Senior. 2016.
\newblock Latent predictor networks for code generation.
\newblock In \emph{Proceedings of the 54th ACL}, pages 599--609, Berlin,
  Germany.

\bibitem[{Luong et~al.(2015)Luong, Pham, and Manning}]{luong-etal:2015:EMNLP}
Thang Luong, Hieu Pham, and Christopher~D. Manning. 2015.
\newblock Effective approaches to attention-based neural machine translation.
\newblock In \emph{Proceedings of the 2015 EMNLP}, pages 1412--1421, Lisbon,
  Portugal.

\bibitem[{Manning et~al.(2014)Manning, Surdeanu, Bauer, Finkel, Bethard, and
  McClosky}]{manning-EtAl:2014:P14-5}
Christopher~D. Manning, Mihai Surdeanu, John Bauer, Jenny Finkel, Steven~J.
  Bethard, and David McClosky. 2014.
\newblock The {Stanford} {CoreNLP} natural language processing toolkit.
\newblock In \emph{ACL System Demonstrations}, pages 55--60.

\bibitem[{Narasimhan et~al.(2016)Narasimhan, Yala, and
  Barzilay}]{narasimhan:2016:EMNLP2016}
Karthik Narasimhan, Adam Yala, and Regina Barzilay. 2016.
\newblock Improving information extraction by acquiring external evidence with
  reinforcement learning.
\newblock In \emph{Proceedings of the 2016 EMNLP}, pages 2355--2365, Austin,
  TX.

\bibitem[{Narayan and Gardent(2014)}]{narayan-gardent:2014}
Shashi Narayan and Claire Gardent. 2014.
\newblock Hybrid simplification using deep semantics and machine translation.
\newblock In \emph{Proceedings of the 52nd ACL}, pages 435--445, Baltimore, MD.

\bibitem[{Papineni et~al.(2002)Papineni, Roukos, Ward, and Zhu}]{papineni:2002}
Kishore Papineni, Salim Roukos, Todd Ward, and Wei-Jing Zhu. 2002.
\newblock Bleu: a method for automatic evaluation of machine translation.
\newblock In \emph{Proceedings of the 40th ACL}, pages 311--318, Philadelphia,
  PA.

\bibitem[{Pascanu et~al.(2013)Pascanu, Mikolov, and Bengio}]{pascanu:2013}
Razvan Pascanu, Tomas Mikolov, and Yoshua Bengio. 2013.
\newblock On the difficulty of training recurrent neural networks.
\newblock \emph{ICML (3)}, 28:1310--1318.

\bibitem[{Pennington et~al.(2014)Pennington, Socher, and
  Manning}]{pennington:2014}
Jeffrey Pennington, Richard Socher, and Christopher~D Manning. 2014.
\newblock Glove: Global vectors for word representation.
\newblock In \emph{In Proceedings of the EMNLP 2014}, volume~14, pages
  1532--43, Doha, Qatar.

\bibitem[{Ranzato et~al.(2016)Ranzato, Chopra, Auli, and
  Zaremba}]{ranzato2016sequence}
Marc’Aurelio Ranzato, Sumit Chopra, Michael Auli, and Wojciech Zaremba. 2016.
\newblock Sequence level training with recurrent neural networks.
\newblock In \emph{Proceedings of ICLR}, San Juan, Puerto Rico.

\bibitem[{Rello et~al.(2013)Rello, Bayarri, G\'{o}rriz, Baeza-Yates, Gupta,
  Kanvinde, Saggion, Bott, Carlini, and Topac}]{Rello:ea:2013}
Luz Rello, Clara Bayarri, Azuki G\'{o}rriz, Ricardo Baeza-Yates, Saurabh Gupta,
  Gaurang Kanvinde, Horacio Saggion, Stefan Bott, Roberto Carlini, and Vasile
  Topac. 2013.
\newblock Dyswebxia 2.0!: More accessible text for people with dyslexia.
\newblock In \emph{Proceedings of the 10th International Cross-Disciplinary
  Conference on Web Accessibility}, pages~--, Brazil.

\bibitem[{Ryang and Abekawa(2012)}]{ryang-abekawa:2012:EMNLP}
Seonggi Ryang and Takeshi Abekawa. 2012.
\newblock Framework of automatic text summarization using reinforcement
  learning.
\newblock In \emph{Proceedings of the 2012 EMNLP-CoNLL}, pages 256--265, Jeju
  Island, Korea.

\bibitem[{Shardlow(2014)}]{Shardlow:2014}
Matthew Shardlow. 2014.
\newblock A survey of automated text simplification.
\newblock \emph{International Journal of Advanced Computer Science and
  Applications}, pages 581--701.
\newblock Special Issue on Natural Language Processing.

\bibitem[{Siddharthan(2004)}]{Siddharthan:06}
Advaith Siddharthan. 2004.
\newblock Syntactic simplification and text cohesion. in research on language
  and computation.
\newblock \emph{Research on Language and Computation}, 4(1):77--109.

\bibitem[{Siddharthan(2014)}]{Siddharthan:2014}
Advaith Siddharthan. 2014.
\newblock A survey of research on text simplification.
\newblock \emph{International Journal of Applied Linguistics}, 165(2):259--298.

\bibitem[{Smith and Eisner(2006)}]{smith2006quasi}
David~A Smith and Jason Eisner. 2006.
\newblock Quasi-synchronous grammars: Alignment by soft projection of syntactic
  dependencies.
\newblock In \emph{Proceedings of the NAACL 206 Workshop on Statistical Machine
  Translation}, pages 23--30, New York City.

\bibitem[{Snover et~al.(2006)Snover, Dorr, Schwartz, Micciulla, and
  Makhoul}]{snover2006study}
Matthew Snover, Bonnie Dorr, Richard Schwartz, Linnea Micciulla, and John
  Makhoul. 2006.
\newblock A study of translation edit rate with targeted human annotation.
\newblock In \emph{Proceedings of association for machine translation in the
  Americas}, volume 200.

\bibitem[{Specia et~al.(2012)Specia, Jauhar, and
  Mihalcea}]{specia-jauhar-mihalcea:2012:STARSEM-SEMEVAL}
Lucia Specia, Sujay~Kumar Jauhar, and Rada Mihalcea. 2012.
\newblock Semeval-2012 task 1: English lexical simplification.
\newblock In \emph{In Proceedings of {*SEM 2012}}, pages 347--355,
  Montr\'{e}al, Canada.

\bibitem[{Sutskever et~al.(2014)Sutskever, Vinyals, and
  Le}]{sutskever2014sequence}
Ilya Sutskever, Oriol Vinyals, and Quoc~V Le. 2014.
\newblock Sequence to sequence learning with neural networks.
\newblock In \emph{Advances in Neural Information Processing Systems}, pages
  3104--3112.

\bibitem[{Vickrey and Koller(2008)}]{vickrey-koller:2008:ACLMain}
D.~Vickrey and D.~Koller. 2008.
\newblock Sentence simplification for semantic role labeling.
\newblock In \emph{Proceedings of ACL-08: HLT}, pages 344--352, Columbus, OH.

\bibitem[{Watanabe et~al.(2009)Watanabe, Junior, {de Uz\~{e}da}, {de Mattos
  Fortes}, Pardo, and Alu\'{ı}sio}]{Watanabe:ea:09}
Willian~Massami Watanabe, Arnaldo~Candido Junior, Vin\'{i}cius~Rodriguez {de
  Uz\~{e}da}, Renata~Pontin {de Mattos Fortes}, Thiago Alexandre~Salgueiro
  Pardo, and Sandra~Maria Alu\'{ı}sio. 2009.
\newblock Facilita: reading assistance for low-literacy readers.
\newblock In \emph{Proceedings of the 27th ACM International Conference on
  Design of Communication}, Bloomington, IN.

\bibitem[{Williams(1992)}]{williams1992simple}
Ronald~J Williams. 1992.
\newblock Simple statistical gradient-following algorithms for connectionist
  reinforcement learning.
\newblock \emph{Machine learning}, 8(3-4):229--256.

\bibitem[{Woodsend and Lapata(2011)}]{woodsend-lapata:2011:EMNLP}
Kristian Woodsend and Mirella Lapata. 2011.
\newblock Learning to simplify sentences with quasi-synchronous grammar and
  integer programming.
\newblock In \emph{Proceedings of the 2011 EMNLP}, pages 409--420, Edinburgh,
  Scotland.

\bibitem[{Woodsend and Lapata(2014)}]{Woodsend:Lapata:2014}
Kristian Woodsend and Mirella Lapata. 2014.
\newblock Text rewriting improves semantic role labeling.
\newblock \emph{Journal of Artificial Intelligence Research}, 51:133--164.

\bibitem[{Wubben et~al.(2012)Wubben, Van Den~Bosch, and
  Krahmer}]{wubben2012sentence}
Sander Wubben, Antal Van Den~Bosch, and Emiel Krahmer. 2012.
\newblock Sentence simplification by monolingual machine translation.
\newblock In \emph{Proceedings of the 50th ACL}, pages 1015--1024, Jeju Island,
  Korea.

\bibitem[{Xu et~al.(2015{\natexlab{a}})Xu, Ba, Kiros, Cho, Courville,
  Salakhutdinov, Zemel, and Bengio}]{xu2015show}
Kelvin Xu, Jimmy Ba, Ryan Kiros, Kyunghyun Cho, Aaron Courville, Ruslan
  Salakhutdinov, Richard~S Zemel, and Yoshua Bengio. 2015{\natexlab{a}}.
\newblock Show, attend and tell: Neural image caption generation with visual
  attention.
\newblock \emph{arXiv preprint arXiv:1502.03044}, 2(3):5.

\bibitem[{Xu et~al.(2015{\natexlab{b}})Xu, Callison-Burch, and
  Napoles}]{Xu-EtAl:2015:TACL}
Wei Xu, Chris Callison-Burch, and Courtney Napoles. 2015{\natexlab{b}}.
\newblock Problems in current text simplification research: New data can help.
\newblock \emph{Transactions of the Association for Computational Linguistics},
  3:283--297.

\bibitem[{Xu et~al.(2016)Xu, Napoles, Pavlick, Chen, and
  Callison-Burch}]{Xu_TACL16}
Wei Xu, Courtney Napoles, Ellie Pavlick, Quanze Chen, and Chris Callison-Burch.
  2016.
\newblock Optimizing statistical machine translation for text simplification.
\newblock \emph{Transactions of the Association for Computational Linguistics},
  4:401--415.

\bibitem[{Yamada and Knight(2001)}]{yamada2001syntax}
Kenji Yamada and Kevin Knight. 2001.
\newblock A syntax-based statistical translation model.
\newblock In \emph{Proceedings of the 39th ACL}, pages 523--530, Toulouse,
  France.

\bibitem[{Zaremba et~al.(2014)Zaremba, Sutskever, and Vinyals}]{zaremba:2014}
Wojciech Zaremba, Ilya Sutskever, and Oriol Vinyals. 2014.
\newblock Recurrent neural network regularization.
\newblock \emph{arXiv preprint arXiv:1409.2329}.

\bibitem[{Zhang and Lapata(2014)}]{zhang-lapata:2014:EMNLP2014}
Xingxing Zhang and Mirella Lapata. 2014.
\newblock Chinese poetry generation with recurrent neural networks.
\newblock In \emph{Proceedings of the 2014 EMNLP}, pages 670--680, Doha, Qatar.

\bibitem[{Zhu et~al.(2010)Zhu, Bernhard, and Gurevych}]{zhu2010monolingual}
Zhemin Zhu, Delphine Bernhard, and Iryna Gurevych. 2010.
\newblock A monolingual tree-based translation model for sentence
  simplification.
\newblock In \emph{Proceedings of the 23rd COLING}, pages 1353--1361, Beijing,
  China.

\end{thebibliography}
\bibliographystyle{emnlp_natbib}

\end{document}